\documentclass{article}

\PassOptionsToPackage{numbers, sort, compress}{natbib}
\usepackage[final]{nips_2016}

\usepackage[utf8]{inputenc} % allow utf-8 input
\usepackage[T1]{fontenc}    % use 8-bit T1 fonts
\usepackage{hyperref}       % hyperlinks
\usepackage{url}            % simple URL typesetting
\usepackage{booktabs}       % professional-quality tables
\usepackage{amsfonts}       % blackboard math symbols
\usepackage{nicefrac}       % compact symbols for 1/2, etc.
\usepackage{microtype}      % microtypography

\title{Clustering with a Reject Option: Interactive Clustering as Bayesian Prior Elicitation}

\author{
  Akash Srivastava\thanks{} \\
  Informatics Forum, University of Edinburgh\\
  10 Crichton St, Edinburgh, Midlothian EH8 9AB, UK \\
  \texttt{akash.srivastava@ed.ac.uk} \\
  \And
  James Zou\\
  Microsoft Research and Stanford University \\
  One Memorial Drive, Cambridge, MA 02142, USA \\
  \texttt{jamesyzou@gmail.com} \\
  \And
  Charles Sutton \\
  Informatics Forum, University of Edinburgh \\
  10 Crichton St, Edinburgh, Midlothian EH8 9AB, UK \\
  \texttt{csutton@inf.ed.ac.uk} \\
}

\usepackage{sutton_math_symbols}
\usepackage{graphicx} 
\usepackage{subcaption} 
\usepackage{amsmath}
\usepackage{xspace}
\usepackage{todonotes}
\usepackage{natbib}
\newcommand{\ciif}{\textsc{Tinder}\xspace}
\newcommand{\jzcomment}[1]{\noindent{\textcolor{blue}{\textbf{\#\#\# JZ:} \textsf{#1} \#\#\#}}}

\newcommand{\cut}[1]{}

\begin{document}

\maketitle

\begin{abstract}
A good clustering can help a data analyst to
explore and understand a data set,
but what constitutes a good clustering may depend
on domain-specific and application-specific
criteria. These criteria can be difficult  to
formalize, even when it is easy for an analyst to know  
a good clustering when she sees one.
We present a new approach to interactive clustering
for data exploration,
called \ciif,
based on a particularly simple feedback mechanism,
in which an analyst can choose to reject individual clusters
and request new ones. The new clusters should be different from previously rejected clusters while still fitting the data well.
We formalize this interaction in a novel Bayesian prior elicitation
framework. In each iteration, the prior is adapted to account for all the previous feedback, and a new clustering is then produced from the posterior distribution.
To achieve the computational efficiency necessary for an interactive setting,
we propose an incremental optimization method over data minibatches using Lagrangian relaxation. Experiments demonstrate that \ciif can produce accurate and diverse clusterings. 
\end{abstract}
\section{Introduction}
\label{introduction}
 
Clustering is a popular tool for exploratory data analysis. A good clustering can help to guide the analyst to better understanding of the data set at hand. An informative clustering captures not only the properties of the data, but also the goals of the analyst. What makes it challenging to identify a good clustering is that it is often difficult to encode the analyst's goals explicitly as machine learning objectives. Moreover, in many settings, the analyst does not have a well-specified objective in mind prior to encountering the data, but rather continuously updates her goals as she learns more through exploratory analysis.  
%The design of a clustering algorithm necessarily reflects prior assumptions about what types of clusters are meaningful. For example, these assumptions manifest in the distance
%metric for a $k$-means clustering algorithm or the choice of the prior distribution
%and likelihood when clustering using a probabilistic model.
%This raises an obvious chicken-and-egg problem: exploration via clustering is a major tool for helping an analyst learn about a data set, but such exploration is likely to influence their opinion about what types of clusters would be meaningful.
%
%Put another way: 
Because the clustering problem is ill-posed, many good clusterings 
of similar quantitative value
exist for a given data set.  Even if a clustering algorithm succeeds in finding a quantitatively good clustering, it still may not be what the user qualitatively wanted.  Nevertheless, the data analyst may not be able to formalize precisely as a quantitative criterion what differentiates a ``good'' clustering from a ``bad'' one.  Still, it seems reasonable to expect that the analyst will know a good clustering when she sees one. 

This gap between formal clustering criteria and the user's exploratory intuition is the motivation for interactive clustering 
\cite{cutting1992scatter,balcan2008clustering,wagstaff2001constrained,bekkerman2007interactive}
and alternative clustering approaches \cite{caruana2006meta,cui2010learning,jain2008simultaneous,dang10cami}. Interactive clustering methods focus
on allowing the user to specify precisely how the clustering should be improved,
such as by splitting or merging clusters \cite{cutting1992scatter,balcan2008clustering}.
Although this can be useful, there are other situations in which
the analyst can tell that a clustering does not meet
her exploratory needs, without having a clear idea of
how it should be improved.
Alternative clustering methods, on the other hand, produce a set of clusterings
which are chosen to be as diverse as possible while still fitting the data.
This supports a more exploratory type of data analysis,
but many such methods do not scale well to an interactive setting,
and sometimes the notion of an alternative is too coarse-grained: An analyst may wish
to preserve some parts of a clustering while discarding others.

To allow the user to provide fine-grained
 ``non-constructive'' feedback on a clustering, we introduce a simple
rejection-based approach to interactive clustering, 
in which the analyst chooses to reject a subset of clusters and replace them
with different ones. This framework contains
alternative clustering as the special case
in which the user rejects all clusters.
The system returns another clustering, 
which is chosen fit the data as well as possible,
while avoiding the creation of any cluster
that is similar to the rejected ones. To reflect the notion of ``rejecting'' a cluster, we call this interaction mechanism \ciif (Technique for INteractive Data Exploration via Rejection).

We formalize this process in a Bayesian framework,
in which we view the interaction procedure as a mechanism
for prior elicitation.  After the user rejects a set of clusters,
we modify the prior distribution over model parameters to severely downweight regions of the parameter space that would lead to clusters that are similar to those previously rejected.
This prior downweighting is achieved through a mutual information criterion,
defined in such a way to prevent the rejection feedback from simply resulting in label permutation.
In interactive settings, it is important that
the response to the user's feedback be produced quickly,
which suggests the use of a stochastic method,
but unfortunately our penalty function does not decompose into a simple sum
over data points.  To surmount this, we propose
an optimization method that introduces
an auxiliary distribution, similar in spirit
to variational
methods, but that follows a Langrangian duality
type argument rather than Jensen's inequality.
The resulting objective function can then be optimized
using a stochastic coordinate descent algorithm over minibatches
of data points,
which we show to be efficient in practice.
%On both text and image data sets, we show that our proposed approach successfully returns a diverse set of clusterings, each of equivalent quality; we show that these clusterings are much more diverse than would be obtained by, e.g., randomized restarts.

\section{Related Work}
\label{related_work}

Previous work on interactive clustering methods 
 exploits various types of user feedback.
One type is  \textit{must-link} and \textit{cannot-link}
constraints between pairs of data points
\cite{wagstaff2001constrained,basu2004active}.
Alternately, a second type of feedback is to request that entire 
clusters be
\textit{split} or \textit{merged}
\cite{cutting1992scatter,balcan2008clustering,balcan2008discriminative}.
A third type of feedback is for 
the analyst to explicitly choose the set of features
to use in the clustering procedure
 \cite{bekkerman2007interactive,dasgupta2010}.
%Additionally, feedback 
%based on individual features requires that the features of the model
%be individually interpretable, such as words in a document.
%Our work, aims at a simple, general type of feedback
%that is applicable across many domains.
Similarly, interactive methods have been proposed for topic
models using must-link / cannot-link
\cite{andrzejewski2009incorporating}, split/merge \cite{Hu:Boyd-Graber:Satinoff-2011,topicmsc}, and feature-level feedback \cite{jagarlamudi2012incorporating}. 
While all three types of feedback
improve clustering quality,
they require that the analyst have a
certain level of knowledge 
about the data set and her information need, which
might not be appropriate for
a highly exploratory analysis. They can also be quite demanding in requiring active guidance from the user.
We are unaware of previous work 
that uses cluster-level accept/reject feedback 
like we do.

In contrast, alternative clustering methods 
\cite{gondek04,bae06,caruana2006meta,jain2008simultaneous,dang10cami,cui2010learning}
focus
 on generating a  set of high-quality clusterings that
 are chosen to be different from each other, which
 the user can select between.
Work in this area has generated
diverse sets of clusters
by randomly reweighting features \cite{caruana2006meta},
 by 
exploring the space of possible clusterings
 using Markov Chain Monte Carlo \cite{cui2010learning},
or by penalizing the objective function to encourage
clusterings to be diverse \cite{gondek04,jain2008simultaneous,dang10cami}.
Our framework for interactive clustering includes
alternative clustering as a special case, bridging
between interactive and alternative clustering.
In particular, the objective function that we propose
recovers the CAMI method \cite{dang10cami}
as a special case in which the user always rejects all clusters, 
and the objective function is optimized jointly over all clusterings,
rather than one clustering at a time in response to user feedback.
Additionally, the optimization method that we propose (Section~\ref{sec:opt}) is different from the previous work
and necessary for obtaining interactive performance.
%Our method handles the problem of clustering using a combined
%approach of interactively generating diverse and accurate clusterings.
%Using a set of a simple interactions, it provides a unified framework
%to explore the space or possible alternate clusterings as well as
%iteratively refine a given clustering by interacting at cluster-level.

%The logarithm of our prior distribution can be viewed as a 
%regularization term that depends on the 
%distribution of the hidden states of a  latent variable model,
%similar in spirit to posterior regularization
%\cite{ganchev2010posterior}. However, this framework does not directly
%apply to our objective function, as it is not clear how the
%regularization term in our method could be written as an expectation
%over a function that is independent of the model parameters.

%We present our method as a mechanism for prior elicitation, 
%which has a history of work in Bayesian statistics \cite{o2006uncertain}.
%In prior elicitation, the
%objective is to capture expert knowledge in form of probabilistic distributions 
%which can then be used in statistical inference.
%We are unaware of previous work that views interactive clustering
%as a form of prior elicitation.

Our work is similar in motivation to  \textit{diverse subset selection},
which is concerned with selecting subsets of 
data from a collection such that the inter-set diversity and the intra-set diversity is maximized,
for example in summarization \cite{gong2014diverse}. 
The application of diverse $k$-best summarization presented in \citet{gillenwater2015}
is a related problem to alternative clustering, if one considers a set of cluster centroids to be a summary of a data set.
Finally, contrastive learning is aimed at fitting a latent variable model so that 
the latent variables explain the difference between one data set from another,
for example, a data set of Chinese news articles versus a dataset of economic articles
\cite{zou2013contrastive}. Our current work is somewhat analogous to this,
in that to support interactive data exploration, we search for different latent
variable explanations of a \emph{single} data set.

\section{Interactive clustering with rejections}

Now we describe our rejection-based framework for interactive clustering.
We begin with an overview of the interaction method.
The data are first clustered according to a standard clustering algorithm.
We present this clustering to the analyst for inspection, for example, by displaying the data points or the features that are most closely associated with each
cluster. 
% Then the analyst has two options: if the clustering meets the information
% need of the analyst, then they can explore the data set accordingly.
% Otherwise the analyst tells the algorithm to reject the clustering
% and present a different one. If the clustering is rejected, 
% we cluster the data again, modifying the objective function for
% the clustering algorithm to penalize clusterings that are similar
% to the previous one. This is to encourage returning a new clustering
% which is as different as possible from the previous one, 
% but that still fits the data well
% according to the quantitative objective function of the original clustering algorithm.
%% AS: mod for NIPS
Then if the clustering does not meet the information
need of the analyst, she can provide  feedback. 
For each cluster, the analyst can either: (a) \textbf{reject} the cluster if it is not relevant
to her information need, (b) \textbf{accept} the cluster if it is relevant, or else
(c) \textbf{do neither}, expressing no opinion about the cluster.
Once this feedback is complete, we cluster the data again, modifying the objective function of
the clustering algorithm to penalize clusters that are similar to rejected clusters,
and to reward clusters that are similar to accepted ones.
This modified objective encourages the algorithm to return a new clustering
that still fits the data well but that respects the user feedback.
The process can be repeated as many times as desired.
We call each iteration of this process a \emph{feedback iteration}.
%That is, the clustering in FI0 is simply the standard clustering
%returned by the clustering algorithm without any feedback, 
%the clustering from FI1 incorporates user feedback from clustering 0, and so on.
%When computing the clustering from the second and subsequent feedback iterations,
%we include penalty terms to encourage the new clustering to be different
%from all previous clusterings that the analyst has seen,
%so that the clusterings do not oscillate.

Now we describe the clustering method used in \ciif.
We formalize the interaction mechanism as a type of Bayesian prior elicitation \cite{o2006uncertain}.
At each feedback iteration $t$, we perform Bayesian
clustering with parameters $\theta$, but with a different
prior $\pi_t(\theta)$ that strongly downweights
parameter vectors that are associated with rejected clusters,
and strongly upweights parameters that are associated with 
accepted clusters.
%% AS: mod for NIPS
We perform clustering using a standard Bayesian mixture model.
Let $x$ denote a single data item, $h \in \{1,\ldots K\}$ be a discrete
latent variable that indicates the cluster membership of $x$, and the vector
$\theta$ denote all of the model parameters, that is, the parameters of the
prior distribution $p(h|\theta)$ over clusters, and the parameters 
of the conditional distribution $p(x|h, \theta)$ of data items given the cluster label.
At feedback iteration $t$, the data is modelled as
\begin{equation}
\label{eq:mdl}
p_t(x, \theta) = \sum_h p(x | h, \theta) p(h | \theta) \pi_t(\theta).
\end{equation}
As the subscripting in \eqref{eq:mdl} suggests, the prior
distribution will change after every feedback iteration
(in a way that we shall discuss in a moment),
but the other parts of the probabilistic model will not.

For computational reasons, we perform
{maximum a posteriori} (MAP) estimation.
Let  $\xB = (x_1 \ldots x_N)$ denote the data, where $x_i$ 
is a single data point,
and $\hB = (h_1 \ldots h_N)$ denote an assignment of cluster labels
to all data points.  Then, at 
each feedback iteration,  MAP estimation
computes the parameter estimate
 $\theta_t= \max_\theta \log p(\theta | \xB)$ and
 a soft cluster assignment
 $p(\hB | \xB, \theta_t)$ over cluster labels.
 (Note that this distribution has the same functional form across all iterations, and the parameter $\theta_t$ could be different in iteration $t$.)  After reviewing the clustering,
the analyst chooses a set of clusters to accept and reject.
Let $A_t \subseteq \{1, \ldots K\}$ be the indices
of the clusters that the user has accepted and $R_t \subseteq \{1, \ldots K\}$
be those the user has rejected. The sets $A_t$ and $R_t$ are disjoint.
Cluster indices that do not appear in $A_t \cup R_t$ are those
clusters for which the analyst has expressed no opinion.

Now we describe how \ciif produces a revised clustering
at feedback iteration $t.$
Following a Bayesian framework, we interpret the user feedback from clusterings $0$ \ldots $t-1$
as an indirect source of information about the analyst's prior beliefs over $\theta,$ 
that she was unable to encode mathematically into the prior distribution. 
Therefore we define
a revised prior distribution $\pi_t(\theta)$ based on all the previous feedback,
which is designed in such a way that 
the resulting clustering, which we denote
$p(\hB | \xB, \theta_t),$  will respect the feedback.
The prior $\pi_t(\theta)$ has the form
 $$\pi_t(\theta) \propto \pi_0(\theta) \prod_{s=0}^{t-1} \exp\{-\beta f_s(\theta,\theta_s)\},$$
 where $f_s$ is a function that measures how well the parameter vector $\theta$
 respects the feedback $(A_s, R_s)$ from iteration $s$ (lower is better). The parameter $\beta$ is a temperature
parameter.
 
For example, consider the case of ``reject all'' feedback,
in which the user has rejected all previous clusters, that is,
$R_s = \{1, \ldots, K \}$ for all $s$. This special
case has been studied in the literature under the name of
 alternative clustering (Section~\ref{related_work}).
In this context, we want $f_s$ to measure the degree
of similarity between the cluster distribution $p(\hB | \xB, \theta)$ and the cluster distribution
$p(\hB | \xB, \theta_s)$ that the user rejected,
so that new parameters $\theta$ which produce
clusters similar to those from $\theta_s$ will have lower probability.
A naive choice for $f_s(\theta,\theta_s)$ would be to use the negative Kullback-Leibler divergence
between the distributions $p(\hB|\xB,\theta)$ and $p(\hB | \xB,\theta_s)$.
However, in the context of clustering, 
this metric suffers from the issue of label switching, i.e., merely permuting the cluster assignments can produce high divergence. 

Instead, we begin by defining a joint distribution
over the individual cluster labels $h$ and $h_s$ that would be assigned
by the current clustering and the previous clustering
to the same data point $x$. This joint distribution is
\begin{equation}
\label{eq:joint}
p_{\theta, \theta_s}(h,h_s,x)=p(h|x, \theta) p(h_s |x, \theta_s)\tilde{p}(x),
\end{equation}
where $\tilde{p}(x) = N^{-1} \sum_i \delta_{x, x_i}$ is the empirical distribution over data points, for the Kronecker delta function $\delta.$
This now defines a bivariate marginal distribution
\begin{align}
p_{\theta,\theta_s}(h,h_s) %&= \frac{1}{N} \sum_{j=1}^N p_{\theta,\theta_s}(h_{sj},h_j,x_j) \nonumber
 &= \frac{1}{N} \sum_{j=1}^N p(h|x_j,\theta)p(h_{s}|x_j, \theta_s) \label{eq:correq}
\end{align}
that measures the overall dependence
between the two different clusterings, marginalizing
out the data.
In other words, $p_{\theta, \theta_s}$ is the joint distribution 
over pairs of cluster labels that results from randomly choosing
a data item $x$, and clustering
it independently according to the distributions
$p(h|x,\theta)$ and $p(h_s | x, \theta_s)$.
 The distribution $p_{\theta, \theta_s}$ also yields marginal distributions
$p_{\theta}(h) = \sum_{h_s} p_{\theta,\theta_s}(h,h_s)$
and $p_{\theta_s}(h_s) = \sum_h p_{\theta,\theta_s}(h,h_s)$
for each of the individual clusterings, which are simply
the prior probabilities of the cluster labels from each
clustering.

Now we can define $f_s$. We begin with the special case
of reject all feedback.  The distribution $p_{\theta, \theta_s}(h,h_s)$
measures the joint distribution between the new clustering and
the previous one at iteration $s$, so to ensure that these
two clusterings are different, we simply minimize their mutual
information.
This yields
\begin{align}
\label{eq:mi_emp}
f_s(\theta, \theta_s) = I(H;H_s) &= \sum_{h=1}^K \sum_{h_s=1}^K p_{\theta,\theta_s}(h,h_s) \log \frac{p_{\theta,\theta_s}(h,h_s)}{p_{\theta}(h)p_{\theta_s}(h_s)}.
\end{align}
% We note that because $f_s(\theta, \theta') \geq 0$,
% we have that $\pi_t$ will be a proper prior
% if $\pi_0$ is. \cs{Sadly this argument no longer holds for accept feedback.}
To handle accept feedback, $f_s$ takes a similar form,
but the sign is flipped for the clusters in $A_s$, so that $f_s$ encourages similarity rather than dissimilarity. More specifically:
\begin{align}
\label{eq:fboth}
f_s(\theta, \theta_s) &= \sum_{h_s \in R_s} \sum_{h=1}^K p_{\theta,\theta_s}(h, h_s)  \log \frac{p_{\theta,\theta_s}(h,h_s)}{p_{\theta}(h)p_{\theta_s}(h_s)} 
 - \sum_{h_s \in A_s} \sum_{h=1}^K p_{\theta,\theta_s}(h, h_s)  \log \frac{p_{\theta,\theta_s}(h,h_s)}{p_{\theta}(h)p_{\theta_s}(h_s)}. 
\end{align}
Note that clusters for which the user has said ``no opinion'' are in neither $A_s$ nor $R_s$,
and therefore such clusters have no effect on $\pi_t$, and hence no effect on the clustering in subsequent feedback
iterations. 
This completes the definition of $\pi_t$. Now, to compute the revised clustering, we perform MAP estimation on \eqref{eq:mdl},
which is equivalent to maximizing
\begin{equation}
\label{eq:model}
L_t(\theta) = \sum_{j=1}^N \log p(x_j | \theta) - \beta \sum_{s=1}^{t-1} f_s(\theta,\theta_s) + \log \pi_0(\theta),
\end{equation}
where $\beta$  
can now be interpreted as a weighting parameter to bring the terms to a common scale. Denote by $\theta_t$ the new MAP parameter estimate,
i.e., $\theta_t = \max_\theta L_t(\theta)$. Then the new
clustering that is displayed to the analyst is based on the soft 
assignment $p(\hB | \xB, \theta_t)$.

\begin{figure}[!tb]
\minipage{0.32\columnwidth}
  \includegraphics[width=\columnwidth]{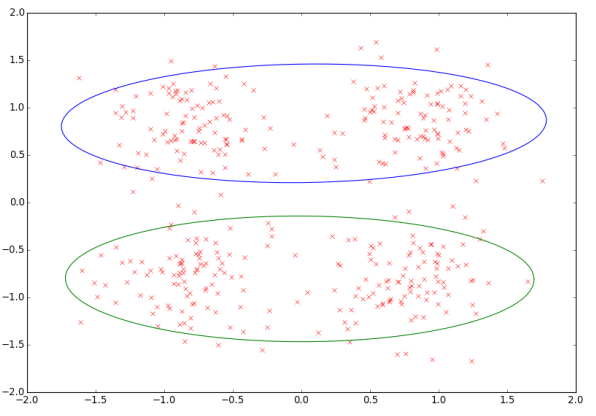}
  (a) Initial clustering
\endminipage\hfill
\minipage{0.32\columnwidth}
  \includegraphics[width=\columnwidth]{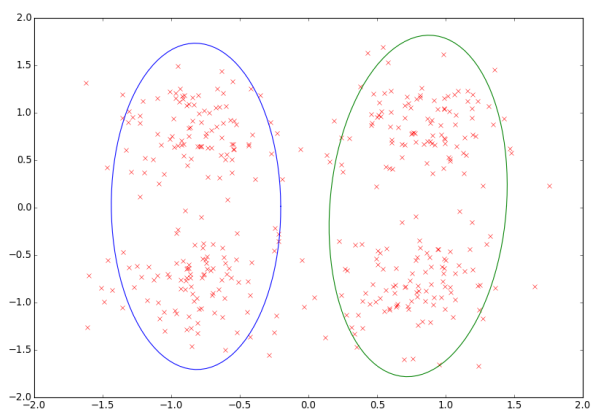}
  (b) After one ``reject all''
\endminipage\hfill
\minipage{0.32\columnwidth}%
  \includegraphics[width=\columnwidth]{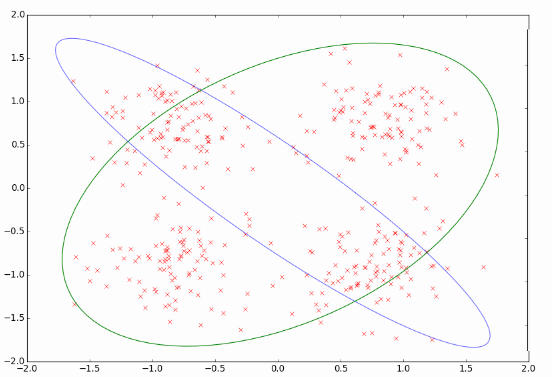}
  (c) After two ``reject all'' 
\endminipage
\caption{Example of \ciif clusterings produced in three feedback iterations on synthetic data, showing (a) the initial clustering 
from expectation maximization (EM), and after (b) one round and (c) two rounds of ``reject all clusters'' feedback
from the user.}
\label{fig:syn}
\end{figure}

\paragraph{Examples.}
As an illustrative example, consider the 2D dataset shown in Figure \ref{fig:syn}(a), which is generated from a mixture of four isometric Gaussians. The ellipses in 
the figure 
show the clustering resulting from maximizing the likelihood of a mixture of two Gaussians using expectation maximization (EM) in the zeroth feedback iteration of \ciif. 
Starting from here, suppose the user rejects both clusters. Then Figure \ref{fig:syn}(b) shows the resulting clustering that \ciif generates in the next feedback iteration. 
Rejecting both clusters again results in the clustering in Figure \ref{fig:syn}(c). Therefore using \ciif, an analyst can obtain three quantitatively different explanations of the data in  three feedback iterations.
 
Our per-cluster feedback framework recovers
alternative clustering, in which the goal is to as explore as many diverse clusterings
as possible, as the special case in which all previous clusters are rejected.
By providing more specific feedback, the user can perform a more directed style of exploration,
in which the user guides the clustering procedure
toward a partitioning that interests her. By incorporating both alternative
clustering and the more directed style of per-cluster feedback in the same framework,
\ciif allows the analyst to flexibly alternate between more exploratory and more directed navigation through the 
space of possible clusterings.

Although we have described \ciif as a clustering method, that is, 
where $p(x,h | \theta)$ is a mixture model, the same logic can be applied
to more general graphical models, e.g., ones that contain other latent
variables in addition to $x$ and $h$. All we require is that the model 
contains a discrete latent variable that we can use in the same way as the cluster labels
$h$, and that MAP estimation of $\theta$ be tractable.
We leave further exploration of this idea to future work.

 \cut{
Figure \ref{fig:mi_contours} provides a deeper insight into \ciif's behavior. It shows the value of $\pi_1(\theta)$ for all possible settings of the cluster centroid of the blue (left) Gaussian if the green (right) Gaussian is held fixed. Notice that the prior alone suggests two optimal settings for the centroid; at the current (shown) location and on top of the green ellipse as both of these settings will minimize the mutual information. Even though the 
mutual information is minimized in case of perfect overlap between the clusters, this is highly discouraged by the likelihood term as it leaves a large part of the data practically unexplained. The joint cost function of \ciif therefore produces clustering (b).\todo{Should we describe the per cluster feedback example figure \ref{fig:mi_contours}(b) as well or there's not enough space? -as}
}

\paragraph{Optimization}
\label{sec:opt}
In this section we discuss how to perform MAP estimation
of $\theta,$ i.e., to efficiently optimize $L_t$.
The gradient of $L_t$ is easy to compute, so it is possible
to apply standard optimization algorithms like conjugate gradient.
However, for an interactive algorithm,
each feedback iteration needs to be relatively fast,
because the computation is run while the user is waiting.
To achieve interactive performance on large data sets, we would therefore prefer a stochastic
gradient style of algorithm, in which each update to
the parameters only depends on a small subset of data points.
But the form of $f_s(\theta, \theta_s)$ makes this difficult.
Notice that the distribution $p_{\theta,\theta_s}(h,h_s)$ contains a summation over data points
within it, and this appears inside a $\log$ within $f_s$.
Therefore the gradient of $L_t$ does not decompose into a simple
sum over data items.

Alternately, we could optimize $L_t$
using the standard EM algorithm for MAP estimation. Recall that
in this algorithm, the E step is unchanged
from maximum likelihood, but the M step contains the log prior
distribution as part of the objective. Applying this to $L_t,$
the M step becomes
$$\theta_t \gets \max_\theta \sum_j \sum_h q_j(h) \log p(h, x_j | \theta) - \beta \sum_s f_s(\theta, \theta_s) + \log \pi_0 (\theta),$$
where $q_j$ is the standard EM auxiliary distribution.
It is not clear that this objective is any easier to optimize
than \eqref{eq:model}, nor is it clear how to derive a stochastic
gradient-style algorithm.

Instead we take a different approach, inspired by Lagrangian relaxation.
To simplify the exposition, we will describe the optimization algorithm only
for the case of ``reject all'' feedback, but the extension to the other types
of feedback is straightforward.
First we introduce an auxiliary random variable $H,$ 
whose output is a cluster assignment, and whose distribution is given
by a variational distribution $q_j(h)$ for each data point $x_j.$
As in \eqref{eq:correq}, we can induce a joint distribution over the random variable $H$  
and the random variable $H_s$ whose distribution
is given by $p(h_{s} | x_j, \theta_s).$ This joint distribution is
$$p_{q,\theta_s}(h, h_s) = N^{-1} \sum_j q_j(h) p(h_{s} | x_j, \theta_s).$$
Notice that this distribution, and therefore the resulting
mutual information, which we denote $I_q(H; H_s),$ is
a function of the variational distribution $q$.
Then optimizing \eqref{eq:model} is equivalent to
\begin{align}
\label{eq:variationalCost}
\max_{\theta,q} &\log p_{\theta}(x)-\beta \sum_s I_q(H;H_s) + \log \pi_0 (\theta).  \\
&\text{s.t.}\, \textnormal{KL}(q_j \,\|\, p(h | \theta, x_j) )=0 \quad \forall j \in \{1,2,\ldots N\} \nonumber,
\end{align}
where KL indicates the Kullback-Leibler divergence.
Incorporating the constraint using a penalty term with parameter $\alpha$ leads to
\begin{equation}
\label{eq:final}
\max_{\theta,q} \log p_{\theta}(x)-\beta \sum_s I_q(H;H_s) - \alpha\sum_j \textnormal{KL}(q_j \,\|\, p(h | \theta, x_j) ) + \log \pi_0(\theta).
\end{equation}
If $\alpha$ is large enough, then the solution of \eqref{eq:final} will be the same as  for \eqref{eq:variationalCost}.
Coordinate descent on \eqref{eq:final} yields the EM-like algorithm:

\noindent 
\textbf{``E''-Step:}
\begin{equation}
\label{e}
q \gets \max_{q} - \beta \sum_s I_q(H;H_s) - \alpha\sum_j \textnormal{KL}(q_j \,\|\, p(h | \theta, x_j) )
\end{equation}
\textbf{``M''-Step:}
\begin{equation}
\label{m}
\theta \gets \max_{\theta} E[\log p(\xB,\hB | \theta)]_{q}
\end{equation}
This is not strictly an EM algorithm, because we lose the lower
bound property that would have arisen if we had applied
Jensen's inequality. 
However, if at the end of optimization
procedure, we have that $q_j(h) = p(h|x_j, \theta)$ for all $j$ (which will happen if $\alpha$ is set high enough, and can be easily checked), then $\theta$ is a local
maximum of $L_t$.

Now we can optimize the objective in the ``E'' step by coordinate descent.
The mutual information $I_q(H; H_s)$ still depends on all of the data points via $p_{q,\theta_s}(h,h_s)$, but now if we perform stochastic
coordinate descent on each distribution $q_j,$ then the value of  $p_{q,\theta_s}(h,h_s)$
can be updated incrementally, so recomputing $I_q(H; H_s)$ does not require iterating
through the entire data set.
The ``M'' step is very fast, as it is exactly the same as 
the M step in the EM algorithm for maximum likelihood.

\section{Experiments}
\label{experiments}
% \subsection{Datasets and Evaluation Metrics}
% \label{dne}

In this section, we evaluate the diversity and the quality
of the clusterings produced by \ciif.
Following previous work in alternative and interactive clustering
\cite{cutting1992scatter,balcan2008clustering,wagstaff2001constrained,bekkerman2007interactive, caruana2006meta,cui2010learning,jain2008simultaneous,dang10cami},
we present an automatic evaluation in which we measure
the quality of clusterings by comparing how well the clusters
correspond to gold standard labels.
An automatic evaluation allows us to compare the output
of the learning algorithms directly without dealing with
difficult and potentially confounding aspects of user interface design.

We evaluate \ciif for both per-cluster feedback (\ciif: Per Cluster), in which
the user feedback is attempting to drive the system toward
a given clustering, and in global mode (\ciif: Global), which is the alternative
clustering setting in which the user is exploring the data set
by rejecting all clusters.  To replicate per-cluster
feedback within an automatic evaluation, we simulate a user
using the following heuristic. 
At each feedback iteration, the user provides feedback on one cluster  at a time. 
If the cluster purity is below 50\% with respect to the gold standard labels,
the simulated user rejects the cluster otherwise the cluster is accepted. If none of the clusters are above this threshold, then the entire clustering is rejected.
The reasoning here is that we are simulating a user whose information
need is to find a clustering similar to that defined by the gold standard labels.

We compare \ciif to the popular \textit{Decorrelated-kMeans} (Dec-kMeans) \cite{jain2008simultaneous} algorithm for alternative clustering,
which uses a penalized $k$-means objective to encourage the centroids
from the previous clustering to be orthogonal to those from the current
clustering.
Since this method in the default setting produces just two clusterings, we extended it by adding additional error and penalty terms to produce more than two clusterings at a time.   
We also compare to running EM using different random initializations,
which will produce different clusterings because 
of its sensitivity to initialization. We call this method \textit{random restarts.}

 We use two data sets:
 a small collection of $640$ face images of $20$ people in different orientations from the CMU face dataset \cite{mitchell1997machine} and a large collection of 10,000 thumbnail images from CIFAR10 \cite{krizhevsky2009learning}.  The CIFAR10 data is significantly
 larger than other data sets that have been used in the alternative
 clustering literature.
 The CIFAR10 data set is labeled with 10 classes.
In the CMU face dataset, each image
has three different types of labels:  the identity of the person in the image, their gender, and the pose (orientation of the face). This provides  three natural clusterings 
of the data. 
% , a medium-sized collection of $1500$ past papers from the proceedings of NIPS \cite{Lichman:2013} 
% For the NIPS dataset we used a simple bag of words representation and 
To obtain features, for the CIFAR10 dataset we use the the embedding generated by training the VGG network \cite{simonyan2014very} on the CIFAR10 training set. For the CMU face dataset, we apply PCA to the raw pixel values and retain 90\% of the variance
from the original data.

To evaluate diversity, the Adjusted Rand Score (ARS) 
is used to measure the distance between two clusterings \cite{hubert1985comparing}; an ARS of $0$ indicates no association
between a pair of clusterings, and a score of $1$ indicates a perfect match between the clusterings. 
To measure the diversity of a set of clusterings, we average
the pairwise ARS over all pairs of clusterings in the set.
To evaluate the quality of a clustering, we report its purity
 with respect to a set of ground truth labels. To evaluate the quality
 of a set of clusterings, we report the maximum purity of any clustering the set,
 reflecting the idea that, after examining a set of different clusterings,
an analyst can choose the single clustering that she finds most useful.
% Given two clusterings, ARS counts the number of pairs assigned to the same clusters and the pairs assigned to different clusters in the two clustering and produces a chance adjusted score between $[-1,1]$ where $-1$ is negative correlation, $0$ is no correlation and $1$ represents a perfect match. Purity for each cluster is calculated by assigning the cluster to the most frequent class and normalizing count of correct prediction by the total size of the dataset.
% \subsection{Experiment Methodology}
% \label{experiment_methodology}
% \subsubsection{Baseline} 

% \subsubsection{Setup}

We use a mixture of Gaussians (GMM) for modeling the CMU Face and the CIFAR10 datasets. In both cases, for the zeroth feedback iteration we set $\pi_0(\theta)$ to be one. 
We reabsorb the relaxed Lagrange multiplier $\alpha$ from Section \ref{sec:opt} in $\beta$ as well. Empirically, we found that TINDER performs well by simply setting $\beta$ such that the penalty term, $\beta \sum_s f$, and the log-likelihood have the same order of magnitude. {Dec-kMeans} also has a similar weighing parameter $\lambda$, which we set according to the guidelines provided by the original authors \cite{jain2008simultaneous}.

All methods are allowed the same  number of feedback iterations,
i.e., all methods are evaluated on the same number of clusterings.
To ensure this, first we run \ciif in per-cluster mode
until the clustering stabilizes, that is, until the simulated
user accepts all clusters. Then we run each of the other methods,
including \ciif Global, for the same number of feedback iterations
that was required by \ciif: Per Cluster.
All methods are repeated $20$ times from different random initializations,
and we report the average maximum quality and the average diversity over the
repetitions. 
For the CMU face data set, we use $K=8$ clusters, whereas for CIFAR10, 
we use $K=10$ clusters.

% \cs{The below doesn't make sense to me. Is what I have
% written accurate: If the user took $k$ feedback iterations before moving on to explore alternative clustering then we ask the other methods (including \ciif Global) to generate $k$ clusterings.
%  This is repeated several times and each time only the clustering with the highest purity is retained and finally averaged to generate the reported purity results. This way each method (including the non-interactive Dec-kMeans) gets a fair chance in terms of the number of clusterings used for evaluation. Once we have generated clusterings of equivalent qualities across all the methods we generate diversity evaluation using all the clusterings generated in the previous process and report them in Table \ref{tab: nmi}. For the CMU dataset we randomly choose to generate eight clusters per clustering whereas we ask for ten clusters in the case of CIFAR10.}

\subsection{Results}
\label{results}
\begin{table}[]
\centering
\caption{Clustering quality, measured by purity to ground
truth labels (higher is better).}
\label{tab:eval}
\begin{tabular}{|l|l|l|l|l|}
\hline
                    & CIFAR10       & \multicolumn{1}{c|}{\begin{tabular}[c]{@{}c@{}}CMU Face\\ Person\end{tabular}} & \multicolumn{1}{c|}{\begin{tabular}[c]{@{}c@{}}CMU Face\\ Gender\end{tabular}} & \multicolumn{1}{c|}{\begin{tabular}[c]{@{}c@{}}CMU Face\\ Pose\end{tabular}} \\ \hline
Random Restarts                  & 0.89          & 0.37                                                                           & 0.87                                                                           & \textbf{0.44}                                                                \\ \hline
Dec-kMeans         & 0.90          & 0.37                                                                           & 0.86                                                                           & 0.42                                                                         \\ \hline
TINDER: Global      & 0.89          & 0.37                                                                           & 0.89                                                                           & 0.40                                                                         \\ \hline
TINDER: Per Cluster & \textbf{0.93} & \textbf{0.39}                                                                  & \textbf{0.93}                                                                  & \textbf{0.44}                                                                \\ \hline
\end{tabular}\vspace{-1ex}
\end{table}

\begin{table}[]
\centering
\caption{Diversity of returned sets of clusterings, measured by Adjusted Rand Score (lower is better).}
% \ciif: Global returns
% a more diverse set of clusterings than other methods, but of equivalent
% quality to the other alternative clustering methods.}

\label{tab: nmi}
\begin{tabular}{|l|l|l|}
\hline
                    & CIFAR10       & CMU Face     \\ \hline
Random Restarts     & 0.56          & 0.55          \\ \hline
Dec-kMeans          & 0.83          & 0.38          \\ \hline
TINDER: Global      & \textbf{0.15} & \textbf{0.27} \\ \hline
TINDER: Per Cluster & 0.88          & 0.59         \\ \hline
\end{tabular}
\end{table}

Table \ref{tab:eval} summarizes the clustering
quality of the methods. The three columns for the CMU Face data report 
quality with respect
to each of the three different types of gold standard labels.
For \ciif: Per Cluster, 
the feedback from the simulated user is based on same set of gold standard labels
that is used to evaluate quality;
that is, the goal is to measure the effectiveness 
of per-cluster feedback at reaching a specific clustering that the user
discovers during exploration.
We find that
\ciif: Per Cluster 
 outperforms or matches the other methods, indicating
that the more specific guidance provided
by per-cluster feedback indeed leads to higher quality clusterings.
On average \ciif requires four feedback iterations to stabilize.
The quality of the clusters from the other three methods
are similar to each other.
 
We report the clustering diversity in Table \ref{tab: nmi}. 
For both the datasets, \ciif: Global clearly returns a much more diverse set of clusterings, outperforming both the baseline methods by a significant margin. 
On CIFAR10, {Dec-kMeans} oscillates between two  similar clusterings and as a result performs worse than {random restarts}. 
These results indicate that overall, \ciif: Global returns a more diverse set of clusters of equivalent quality to the other alternative clustering methods.
As expected, \ciif: Per Cluster results are not as diverse,
because the goal of per-cluster feedback is to drive the method
towards a specific target clustering, which necessarily reduces diversity.
% For CIFAR10, \textit{Dec-kMeans} fails to explore much even in comparison to the \textit{random restarts} methods. Though for the faces dataset it does better than \textit{random restarts} but does not reach close to \ciif Global.   
\begin{figure}[!tb]
\minipage{0.48\linewidth}
  \centerline{\includegraphics[width=\columnwidth]{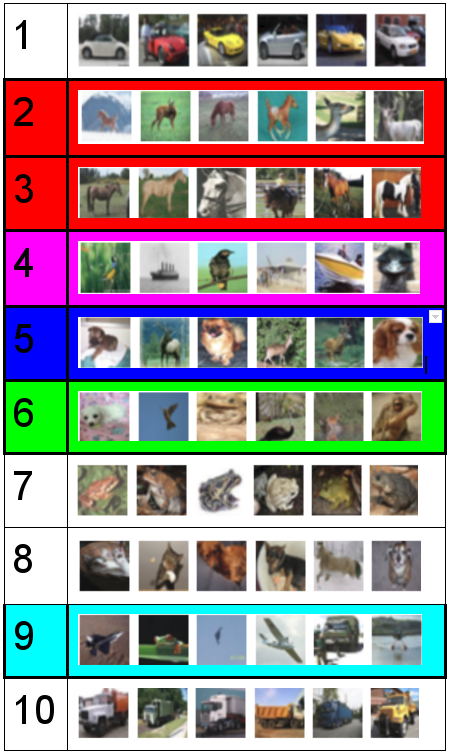}}
  \subcaption{Clustering 0 (no feedback)}
\endminipage\hfill
\minipage{0.48\columnwidth}
  \centerline{\includegraphics[width=\columnwidth]{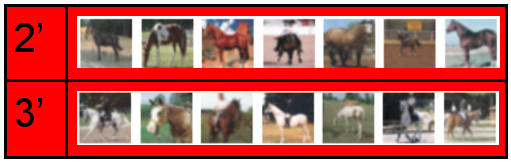}}
  \subcaption{Clustering 1 (after one feedback iteration). For space, only
  two of the ten clusters are shown.}\vspace{2ex}
  \centerline{\includegraphics[width=\columnwidth]{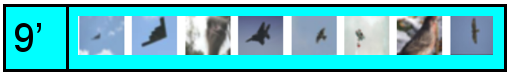}}
  \subcaption{Clustering 2 (after two feedback iterations). For space, only
  one of the ten clusters is shown.}\vspace{2ex}
  \centerline{\includegraphics[width=\columnwidth]{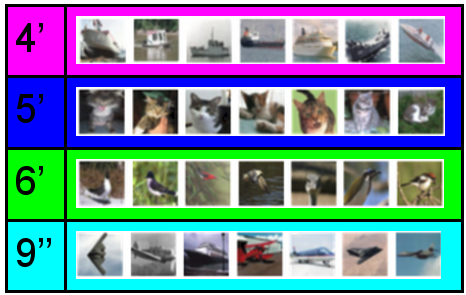}}
  \subcaption{Clustering 5 (after five feedback iterations). For space, only
  four of the ten clusters are shown.}
\endminipage\hfill
\caption{Example of \ciif clusterings on the CIFAR10 dataset.}\vspace{-1ex}\label{fig:cifarex}
% \jzcomment{This is better but still not clear in (b) it's cluster 8 that splits right? Move the horse cluster in (a) to the top, circle it and have two arrows pointing out the splits. Also label that this is the first FI. In (c) I don't know what's the cluster 11 + 14 you are citing. Show these in (a). Make it so that they are next to (c). (d) it'd be more striking if you show the cluster in (a) that were replaced by these. }
\end{figure}

To illustrate the effect of the feedback, we display in  Figure \ref{fig:cifarex}
some of the clusters from \ciif: Global on the CIFAR10 dataset. \ciif: Global clusterings are not just able to find all the original CIFAR10 clusters but other meaningful clusters as well. 
In the figure, each of the rows represents a cluster and shows the top 6 images from that cluster ordered by their likelihood under the cluster. Figure \ref{fig:cifarex}(a) shows the initial clustering for $K=10$
with no feedback. 
Clustering 1 (Figure \ref{fig:cifarex}(b)) is produced by \ciif  
after a single iteration of ``reject all'' feedback.
We see that Clusters 2 and 3 from Clustering 0
(which contain deer and horses, respectively)
are replaced in Clustering 1 by clusters 2' and 3', which contain 
large animals (Cluster 2') and horses with riders (Cluster 3'). % Neither clusters 2' or 3' Note that neither of these clusters are labeled in the original CIFAR10 categories. 
The result of the next feedback iteration is shown in Clustering 2
(Figure \ref{fig:cifarex}(c)). We see that Cluster 9 has been replaced
by Cluster 9', which contains images of birds and planes, 
which were scattered over multiple clusters in Clustering 0. Finally, after five feedback iterations,
Clustering 5 (Figure \ref{fig:cifarex}(d)) includes clusters of ships (Cluster 4'), cats (Cluster 5'), birds (Cluster 6') and planes (Cluster 9''),
which did not exist in Clustering 0. These new clusters 
replace Clusters 4-6 and 9 from Clustering 0, which have low purity.

As for running time, each feedback iteration of \ciif requires
a few seconds for the CMU Face data set and under a minute for CIFAR10. Our implementation of Dec-kMeans performs comparably. Both methods take the same amount of time as standard EM without feedback.
Finally, we observe that \ciif can be easily applied to any mixture model, not just a  mixture of Gaussians.
To demonstrate this, we also applied \ciif to a mixture of multinomials model for text data (see supplementary  
material).

\section{Conclusion}
\label{conclusion}
In this paper we have presented a method for
interactive clustering based on a particularly
simple feedback mechanism, in which an analyst can 
reject individual clusters and request new
ones.
The interaction is formalized as a method of prior elicitation in a Bayesian model of clustering. 
We showed the efficacy of this method on two real world datasets. An interesting direction of future work would be to extend our approach to other graphical models for data exploration, such as topic models.

%cs: Remove for blind review
%\subsubsection*{Acknowledgments}

{\small\setlength{\bibsep}{2pt}
\bibliography{example_paper}
}

\end{document}